\documentclass[12pt]{spieman} 
\usepackage{amsmath,amsfonts,amssymb}
\usepackage{graphicx}
\usepackage{setspace}
\usepackage{tocloft}
\usepackage{textcomp}
\usepackage{color,soul}
\usepackage{subcaption}
\usepackage{float}
\usepackage[colorlinks=true, allcolors=blue]{hyperref}
\usepackage[table,xcdraw,dvipsnames]{xcolor}
\usepackage{lineno}
\usepackage{array}
\usepackage{listings}
\usepackage{mathptmx}
\usepackage{epsfig}
\usepackage{rotating}
\usepackage{makeidx}
\usepackage{url}
\usepackage{booktabs}
\usepackage{tabularx}
\usepackage{blindtext}
\usepackage{times}
\usepackage{nccmath}
\usepackage{mwe}
\usepackage{acro}
\usepackage{adjustbox}
\usepackage{colortbl}
\usepackage{relsize}
\usepackage{pifont}
\usepackage{multirow}
\usepackage{multicol}
\usepackage{nicematrix}
\usepackage{bm}
\usepackage{makecell}
\usepackage{tabu}
\usepackage[capitalize]{cleveref}
\DeclareMathSymbol{\shortminus}{\mathbin}{AMSa}{"39}

\usepackage{color}
\definecolor{dkgreen}{rgb}{0,0.6,0}
\definecolor{gray}{rgb}{0.5,0.5,0.5}
\definecolor{mauve}{rgb}{0.58,0,0.82}

\colorlet{Mycolor1}{green!10!orange}

\usepackage{lineno}
\modulolinenumbers[1]

\title{Evaluating Cell AI Foundation Models in Kidney Pathology with Human-in-the-Loop Enrichment}

\author[a]{Junlin Guo}
\author[a]{Siqi Lu}
\author[b]{Can Cui}
\author[b]{Ruining Deng}
\author[b]{Tianyuan Yao}
\author[b]{Zhewen Tao}
\author[c]{Yizhe Lin}
\author[b]{Marilyn Lionts}
\author[b]{Quan Liu}
\author[a]{Juming Xiong}
\author[e]{Yu Wang}
\author[e]{Shilin Zhao}
\author[a,b]{Catie Chang}
\author[a]{Mitchell Wilkes}
\author[d]{Agnes Fogo}
\author[d,f]{Mengmeng Yin}
\author[d]{Haichun Yang}
\author[a,b,d]{Yuankai Huo}

\affil[a]{Department of Electrical and Computer Engineering, Vanderbilt University, Nashville, TN, USA}
\affil[b]{Department of Computer Science, Vanderbilt University, Nashville, TN, USA}
\affil[c]{Department of Mathematics, Vanderbilt University, Nashville, TN, USA}
\affil[d]{Department of Pathology, Microbiology and Immunology, Vanderbilt University Medical Center, Nashville, TN, USA}
\affil[e]{Department of Biostatistics, Vanderbilt University Medical Center, Nashville, TN, USA}
\affil[f]{Shanghai Ninth People’s Hospital affiliated to Shanghai JiaoTong University School of Medicine, Shanghai, China}

\cftpagenumbersoff{figure}
\cftpagenumbersoff{table} 
\begin{document} 
\begin{sloppypar}
\maketitle

\begin{abstract}

\textbf{Background}
Large-scale artificial intelligence foundation models have emerged as promising tools for addressing healthcare challenges, including digital pathology. While many have been developed for complex tasks such as disease diagnosis and tissue quantification using extensive and diverse datasets, their readiness for seemingly simpler tasks, such as nuclei segmentation within a single organ (for example, the kidney), remains unclear. This study answers two questions: How good are current cell foundation models? and How can we improve them?
\\
\textbf{Methods}
We curated a multi-center, multi-disease, and multi-species dataset sampled from 2,542 kidney whole slide images. Three state-of-the-art cell foundation models—Cellpose, StarDist, and CellViT—were evaluated. To enhance performance, we developed a human-in-the-loop strategy that distilled multi-model predictions, improving data quality while reducing reliance on pixel-level annotation. Fine-tuning was performed using the enriched datasets, and segmentation performance was quantitatively assessed.
\\
\textbf{Results}
Here we show that cell nuclei segmentation in kidney pathology still requires improvement with more organ-targeted foundation models. Among the evaluated models, CellViT achieves the highest baseline performance, with an F1 score of 0.78. Fine-tuning with enriched data improves all three models, with StarDist achieving the highest F1 score of 0.82. The combination of the foundation model–generated pseudo-labels and a subset of pathologist-corrected ``hard” patches yields consistent performance gains across all models.
\\
\textbf{Conclusions}
This study establishes a benchmark for the development and deployment of cell AI foundation models tailored to real-world data. The proposed framework, which leverages foundation models with reduced expert annotation, supports more efficient workflows in clinical pathology.

\end{abstract}

\textbf{Plain language summary}
\\
The rise of digital pathology has transformed traditional histology slides into vast collections of high-resolution images, enabling medical research on a much larger scale. However, analysing this data remains challenging. Foundation models—advanced AI systems trained on diverse datasets—offer a promising solution, but their ability to perform simpler yet essential tasks, such as identifying cell nuclei in kidney tissue, is unclear. We evaluated three leading models on a large, curated kidney image dataset and found that cell nuclei segmentation in kidney pathology still requires improvement with more organ-targeted foundation models. To enhance performance, we introduced a ``human-in-the-loop'' approach that combines multiple foundation models with expert labeling of only the most difficult cases, improving accuracy, reducing manual labeling, and enabling more efficient pathology workflows.

\begin{spacing}{2}  


\section{Introduction}

AI foundation models trained on massive, diverse datasets are widely applied across numerous fields, including healthcare~\cite{moor2023foundation, bommasani2021opportunities}. Their versatility enables these models to tackle a variety of downstream tasks, with one of the most prominent applications being digital pathology~\cite{ma2024segment, deng2023segment,mazurowski2023segment,cui2024pfps, wu2023medical}. In this field, cell instance segmentation serves as a critical first step for extracting biologically meaningful information used in diagnosis, treatment planning, and research~\cite{image_based_cell_profiling, deep_learning_pathology_survey, whole_cell_segmentation, structured_tumor_immune, medical_image_analysis_survey, image_based_cell_phenotyping}. Accurate cell or nuclei segmentation underpins downstream tasks such as cell type classification~\cite{liu2019acute}, cell counting~\cite{tran2018blood}, spatial transcriptomics~\cite{zhu2025asign}, and phenotype analysis~\cite{bougen2017large}, and is essential for whole slide image (WSI) analysis~\cite{nasir2023nuclei}.

The task of segmenting nuclei has evolved considerably over the past two decades. Early solutions relied on image processing techniques such as marker-controlled watershed segmentation~\cite{yang2006nuclei, cheng2008segmentation}, active contour models~\cite{ali2012integrated}, and multi-pass watershed methods~\cite{tareef2018multi}. While effective in specific contexts, these methods depended on hand-crafted features and were sensitive to parameter tuning, making them less robust for heterogeneous or overlapping nuclei. The advent of deep learning (DL) addressed many of these limitations by replacing manual feature design with learned representations. Two-stage DL approaches, such as Mask R-CNN~\cite{he2017mask}, first localize nuclei before refining their boundaries, whereas one-stage methods integrate detection and segmentation into a unified process. Notable one-stage models include Micro-Net~\cite{raza2019micro}, HoVer-Net~\cite{hover-net}, StarDist~\cite{stardist}, CPP-Net~\cite{chen2023cpp}, Cellpose~\cite{cellposs}, and CellViT~\cite{cellvit}, which vary in architectural design but share the advantages of efficiency and standardization.

More recently, foundation models for cell segmentation~\cite{cellposs, stardist, cellvit, israel2023foundation} have emerged, which are trained on large and diverse datasets encompassing multiple cell types, microscopy techniques, and experimental conditions. These models have demonstrated impressive cross-domain generalization, enabling strong performance in unfamiliar settings. However, despite their success in complex tasks such as disease diagnosis and tissue quantification, it remains unclear whether they can generalize to simpler yet essential tasks, such as cell nuclei segmentation within a single organ. This is a relevant and underexplored question in kidney pathology, where WSIs feature high cellular diversity—at least 16 specialized epithelial types along with endothelial, immune, and interstitial cells~\cite{balzer2022many}—and where common stains such as Periodic Acid–Schiff (PAS) are rarely represented in training datasets.

Beyond large-scale training and model architecture, the availability of high-quality training data remains a critical bottleneck. Producing pixel-level annotations for instance segmentation is time-consuming and expensive (typically requiring $10^{-2} \text{ to } 10^{0}$ hours per image)~\cite{whole_cell_segmentation, van2016deep, Schwartz803205}. Human-in-the-loop (HITL) strategies can mitigate this burden by iteratively refining model predictions with expert feedback~\cite{pachitariu2022cellpose, whole_cell_segmentation, Schwartz803205}. For example, Cellpose 2.0~\cite{pachitariu2022cellpose} fine-tunes pretrained models using as few as 100–200 region-of-interest annotations. While effective, most HITL pipelines are built around a single model, leaving the potential benefits of combining multiple foundation models largely unexplored.

In this study, we address these gaps by evaluating three widely used cell AI foundation models—Cellpose~\cite{cellposs}, StarDist~\cite{stardist}, and CellViT~\cite{cellvit}—on a diverse evaluation dataset that includes kidney nuclei data sampled from 2,542 kidney WSIs sourced from humans and rodents across both public and in-house datasets. To our knowledge, the scale of this study's kidney WSIs surpasses all publicly available labeled nuclei datasets that include the kidney. Here we show that cell nuclei segmentation in kidney pathology still requires improvement with more organ-targeted models. To enhance performance, we introduce a HITL data enrichment strategy that combines foundation model predictions with expert corrections for the most challenging cases. Fine-tuning with these enriched datasets improves all three models, with StarDist achieving an F1 score of 0.82. Pairing pseudo-labels with a subset of pathologist-corrected ``hard” patches consistently boosts performance, improves segmentation accuracy, and streamlines workflows in research and clinical pathology.
\begin{figure*} [ht]
\begin{center}
\includegraphics[width=1\linewidth]{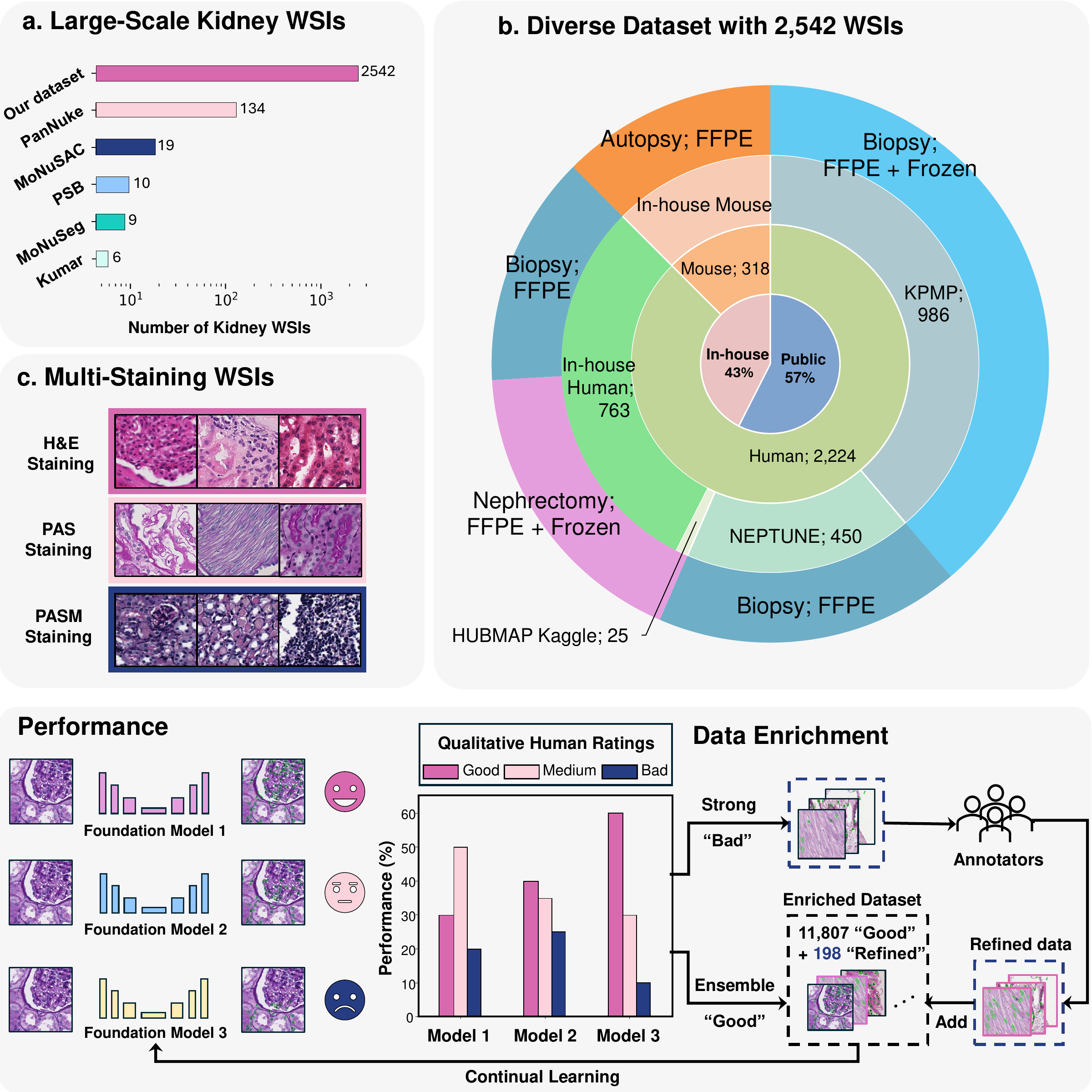}
\end{center}
\caption{\textbf{Overall framework.} The upper panel illustrates the diverse evaluation dataset consisting of 2,542 kidney WSIs. \textbf{Performance:} Kidney cell nuclei instance segmentation was performed using three SOTA cell foundation models: Cellpose, StarDist, and CellViT. Model performance was evaluated based on qualitative human feedback for each prediction mask. \textbf{Data Enrichment:} A human-in-the-loop (HITL) design integrates prediction masks from performance evaluation into the model’s continual learning process, reducing reliance on pixel-level human annotation.}
{ \label{fig:fig1}}
\end{figure*}

\section{Methods}
In this section, we first introduce the overall framework of the study, followed by details of its design. Specifically, we describe the construction of a diverse, large-scale dataset for evaluating foundation models, the human-in-the-loop strategy for enriching the training data by leveraging these models, the continual fine-tuning of models with the enriched dataset, and the experimental setup.

\subsection{Overall Framework}
Fig.~\ref{fig:fig1} illustrates the overall framework of this study. First, we curated a diverse, large-scale evaluation dataset comprising samples from 2,542 kidney whole slide images (Fig.~\ref{fig:fig1}a-c). Next, kidney cell nuclei instance segmentation was performed using three state-of-the-art foundation models: Cellpose~\cite{cellposs}, StarDist~\cite{stardist}, and CellViT~\cite{cellvit}. Model performance was then assessed, as shown in the bottom panel, through qualitative human evaluation of each prediction mask. Lastly, to address the more challenging task of improving model performance, the Fig.~\ref{fig:fig1} Data Enrichment stage employs a human-in-the-loop framework that integrates foundation model predictions with expert corrections for the most difficult cases into the model’s continual learning process, thereby reducing reliance on pixel-level human annotation.

\subsection{Curating a Diverse Large-Scale Dataset}
To provide a comprehensive assessment of their performance in segmenting kidney nuclei, we constructed a diverse evaluation dataset. Our dataset comprises both public and private kidney data, with a total of 2,542 whole-slide images (WSIs). As illustrated in Fig.\ref{fig:fig1}b, 57\% (1,449 WSIs) come from publicly available sources, including the Kidney Precision Medicine Project (KPMP)\cite{kpmp_data}, NEPTUNE~\cite{barisoni2013digital}, and HUBMAP~\cite{hubmap-kidney-segmentation,hubmap2019human}, while the remaining 43\% (1,093 WSIs) originate from an internal collection at Vanderbilt University Medical Center. To increase the diversity of our dataset, we incorporated WSIs from both human and rodent samples. As depicted in Fig.~\ref{fig:fig1}c, these WSIs were stained using Hematoxylin and Eosin (H\&E), Periodic acid-Schiff methenamine (PASM), and Periodic acid-Schiff (PAS), with PAS being widely used in kidney pathology but less frequently in other organs. We randomly extracted four 512$\times$512-pixel image patches from each kidney WSI at 40$\times$ magnification. Patches that were contaminated, of low imaging quality, from dead tissue, or with incorrect staining methods were discarded. This process resulted in an evaluation dataset of 8,818 image patches.

\begin{figure*} [t]
\begin{center}
\includegraphics[width=1\linewidth]{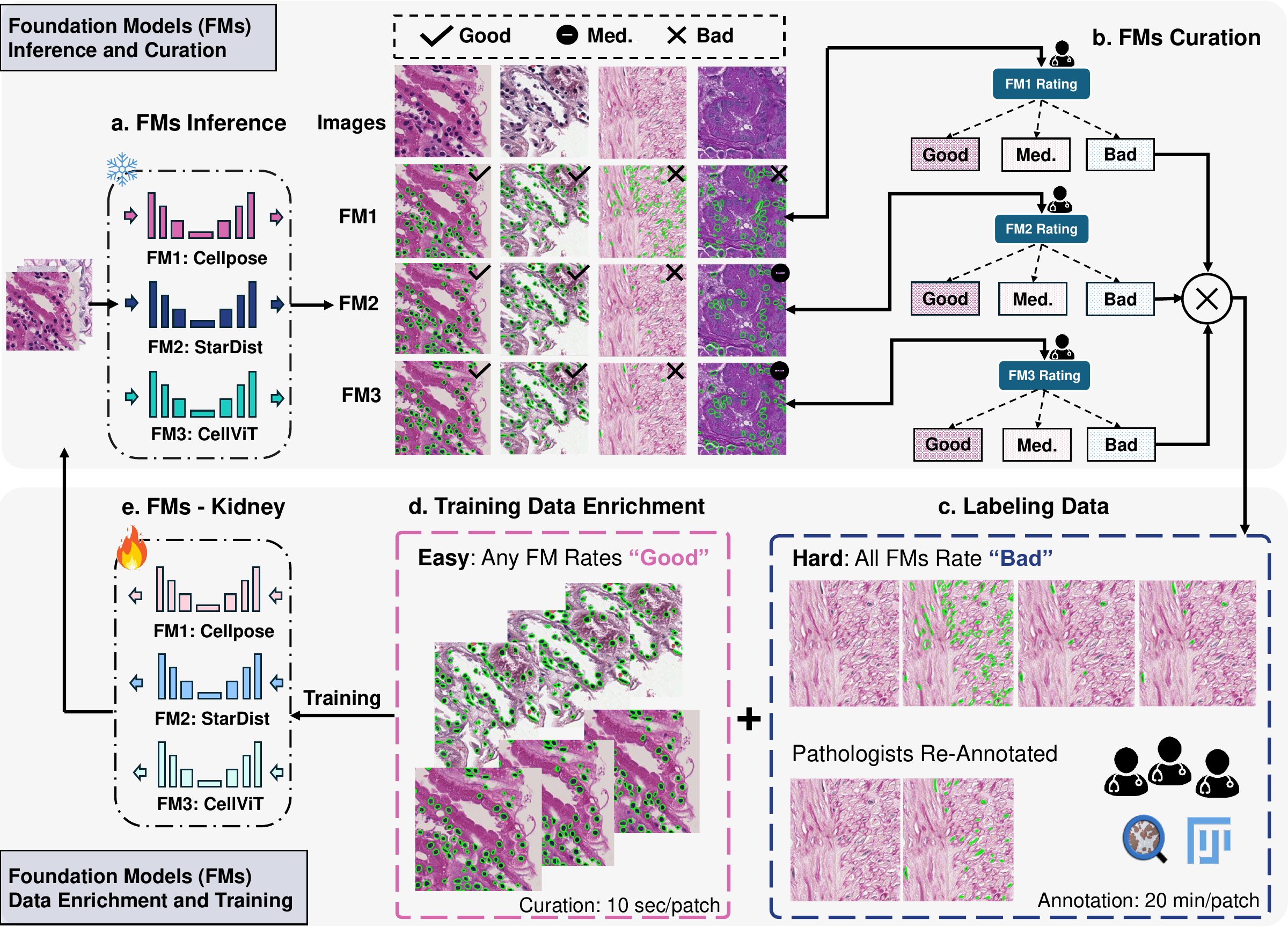}
\end{center}
\caption{\textbf{Human-in-the-loop (HITL) Data Enrichment Design.} The upper panel shows the inference and curation of prediction masks from three foundation models (Cellpose, StarDist, and CellViT) in this study.  \textbf{(a)} First, kidney nuclei instance segmentation was performed on the evaluation dataset using three cell foundation models. \textbf{(b)} Model performance was evaluated by rating each prediction mask as ``good," ``medium," or ``bad" according to criteria from a renal pathologist. ``Good" predictions captured approximately 90\% of the nuclei in a patch, ``bad" predictions captured less than 50\%, and the rest were classified as ``medium." We used this rating system to both qualitatively and quantitatively evaluate and categorize each model’s predictions within our dataset. \textbf{(c-e)} The lower panel illustrates the data enrichment strategy that utilizes these curation outcomes to enhance model performance through continuous fine-tuning. Specifically, to enrich the training dataset while minimizing pixel-level annotation, we used both pseudo-labeled images from multiple foundation models (termed as \textbf{``easy"}) and \textbf{“hard”} samples that all models failed.}
{ \label{fig:HITL}}
\end{figure*}

\subsection{Data Enrichment with Multiple Foundation Models and HITL}

\begin{figure*} [ht]
\begin{center}
\includegraphics[width=1\linewidth]{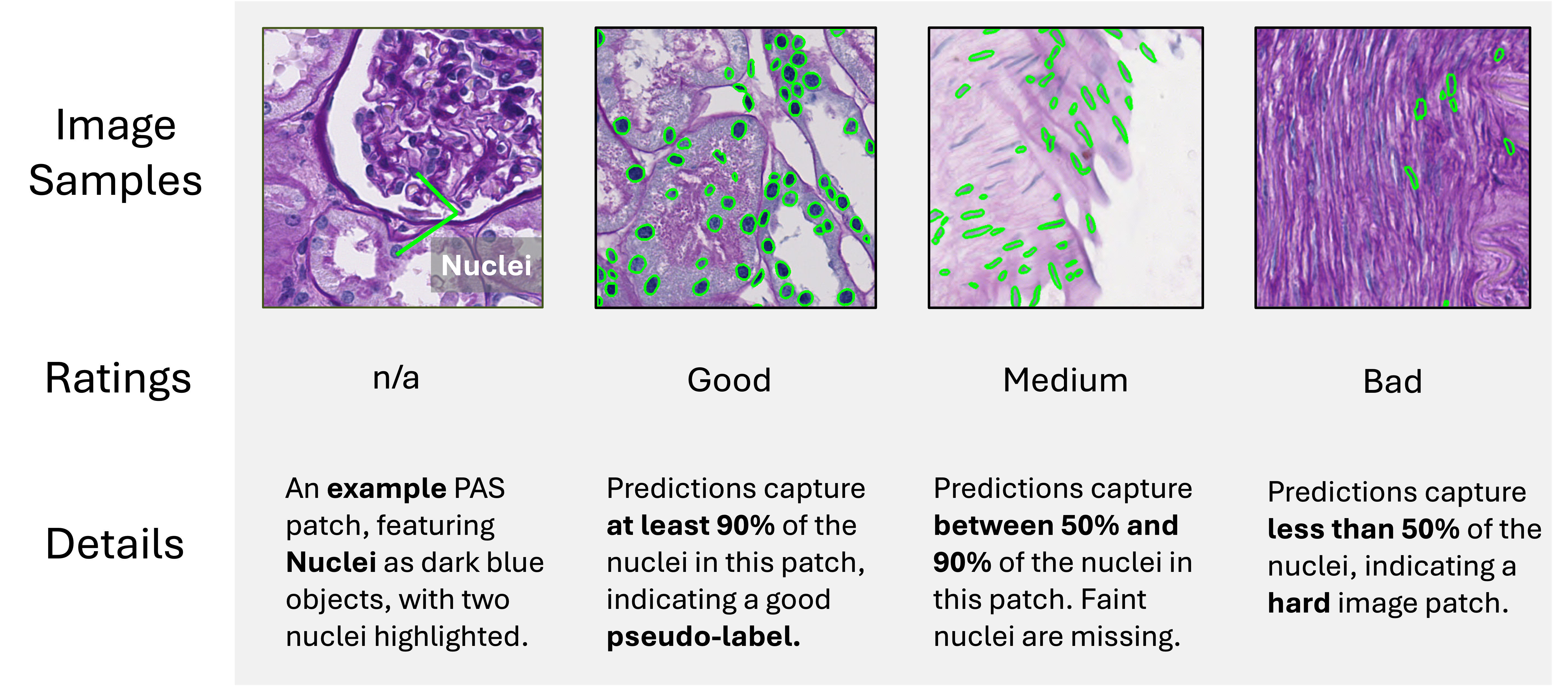}
\end{center}
\caption{\textcolor{black}{\textbf{Illustrations of Rating Criteria.} This Figure illustrates the rating criteria, showing examples and their ratings for ``good”, ``medium”, and ``bad” categories.}} 
{ \label{fig:rating_details}}
\end{figure*}

\textcolor{black}{\textbf{Rating Image Patches.} As shown in Fig.~\ref{fig:HITL}a, nuclei instance segmentation was performed on these kidney nuclei image patches using three cell foundation models: Cellpose~\cite{cellposs}, StarDist~\cite{stardist}, and CellViT~\cite{cellvit}. Then, instead of directly correcting their predicted instance masks, we use a rating-based system to evaluate the predictions from the three foundation models (as shown in Fig.~\ref{fig:HITL}b). The ratings, conducted in a blinded experiment, were performed separately by two pathologist-trained students evaluating model predictions. An experienced renal pathologist subsequently reviewed and validated the samples where the students’ assessments disagreed. As shown in Fig.~\ref{fig:rating_details}, ratings—categorized as ``good,” ``medium,” or ``bad” based on the criteria set by an expert renal pathologist—defined “good” predictions as capturing approximately 90\% of nuclei in a patch, ``bad” as capturing less than 50\%, and all others as ``medium.”}  

\textbf{Pseudo-Labels and Hard Data Annotation.} In this work, we leveraged the image patch curation results to reduce the pixel-level labeling effort in the HITL design. Specifically, we scaled up the training dataset by directly utilizing the ``easy" (pseudo-labeled) samples (shown as Fig.~\ref{fig:HITL}d). To bridge the domain gap, we also incorporated a small set of representative ``hard" samples (shown as Fig.~\ref{fig:HITL}c), manually annotated by pathologists at a cost of roughly 20 minutes per patch. 

\subsection{Data-Enriched Foundation Models Fine-Tuning}
Leveraging curated data from multiple foundation models, we fine-tune them with “easy” patches (pseudo-labels), “hard” patches (pathologist-corrected), or a combination of both to improve kidney-nuclei segmentation. In this section, we (1) introduce the three foundation models used for image-patch curation and continuous fine-tuning, and (2) address a key fine-tuning challenge: the imbalance between ``easy" and ``hard" image patches. To mitigate this imbalance, we apply a customized weighted-oversampling method.

\subsubsection{Cell AI Foundation Models}
This section outlines the three cell foundation models used in this study. Details for each foundation model can be found in Cellpose~\cite{cellposs}, StarDist~\cite{stardist}, and CellViT~\cite{cellvit}. 

\textbf{Cellpose}: \textcolor{black}{Cellpose \cite{cellposs} is a generalist segmentation model that utilizes a U-Net backbone to predict the horizontal and vertical gradients of topological maps, as well as a binary map. Then, a gradient tracking algorithm is employed to enable precise delineation of cellular boundaries. According to~\cite{cellposs}, Cellpose was trained on highly varied images of cells, containing over 70,000 segmented objects, capturing a wide range of imaging modalities and cellular morphologies. The training dataset includes fluorescent microscopy with cytoplasmic markers and DAPI-stained nuclei, brightfield microscopy, membrane-labeled cells,  as well as non-cellular and atypical cellular images from Google search to enhance the model's generalization beyond conventional biological images. Morphologically, it featured diverse cell types, including for example, neurons with complex dendritic processes, bone marrow-derived macrophages, cultured Drosophila and mouse cortical cells, pancreatic stem cells, cancer cell lines such as U2OS osteosarcoma cells, and histological samples from various human organs (e.g., MoNuSeg~\cite{MoNuSeg}). This comprehensive diversity in both imaging types and cell morphologies increases its segmentation ability across a wide range of biological and imaging contexts without requiring retraining or parameter adjustments.}

\textbf{StarDist}: StarDist \cite{stardist} employs a unique star-convex polygon representation for nuclei segmentation, which is particularly effective for roundish objects such as cell nuclei. The model is built on a U-Net backbone and predicts object probabilities and radial distances from the center of each nucleus to its boundary, forming star-convex polygons. StarDist was originally developed for fluorescence microscopy images. To enhance its applicability to histopathology images, the model was further trained on the Lizard dataset~\cite{lizard}, which includes 4,981 histopathology images, each of size 256 $\times$ 256 $\times$ 3, containing six different cell types (neutrophil, epithelial, lymphocyte, plasma, eosinophil, and connective tissue cells). During the post-processing stage, non-maximum suppression (NMS) is used to remove redundant polygons, ensuring that only unique instances are retained. Additionally, test-time augmentations and model ensembling were employed to further enhance segmentation performance, making StarDist a robust solution for diverse microscopy image types.

\textbf{CellViT}: CellViT \cite{cellvit} employs a U-Net shaped hierarchical encoder-decoder Vision Transformer (ViT) \cite{} backbone, designed specifically for nuclei segmentation and classification in histopathology images. The encoder utilizes pre-trained weights from a ViT trained on 104 million histological images (ViT\textsubscript{256})\cite{chen2022scaling}, a model that demonstrated superior performance in cancer subtyping and survival prediction tasks. CellViT is further trained on the PanNuke dataset~\cite{pannuke}, which includes 189,744 annotated nuclei across 19 tissue types, grouped into five clinically relevant classes: neoplastic, inflammatory, epithelial, dead, and connective. The images in this dataset are captured at 40$\times$ magnification with a resolution of 0.25 µm/pixel, making it a challenging benchmark due to its diversity and class imbalance. CellViT's post-processing pipeline follows the HoVer-Net methodology \cite{hover-net}, using gradient maps of horizontal and vertical distances for accurate boundary delineation. Additionally, the network benefits from extensive data augmentation techniques and transfer learning strategies to overcome the scarcity of annotated medical data, achieving SOTA performance on both the PanNuke and MoNuSeg datasets.

\subsubsection{Class-wise Weighted Oversampling}
To mitigate the severe imbalance between ``easy” and ``hard” patches ($\approx$ 50{:}1), we adopt the customized weighted-oversampling strategy described in~\cite{cellvit}. By controlling an oversampling factor \( \gamma_s \in [0, 1] \), higher sample weights will be assigned to underrepresented class (``hard"), ensuring they are sampled more frequently during training. Thus, challenging nuclei samples are emphasized while catastrophic forgetting is reduced by retaining large-scale ``easy" labels. A detailed derivation of the weighted oversampling method is provided in the Supplementary Methods.

\subsection{Experimental Setup}

\subsubsection{Dataset}
\textbf{Evaluation Dataset.} As described in the previous section, the evaluation dataset in this study is highly diverse, comprising 2,542 whole-slide images (WSIs) from both public and private sources. It includes samples from both human and rodent tissue, stained with H\&E, PASM, and PAS. After discarding contaminated or poor-quality patches, we randomly extracted four 512 $\times$ 512-pixel patches from each WSI at 40$\times$ magnification, resulting in an evaluation dataset of 8,818 image patches. As illustrated in Fig.\ref{fig:fig1}b, 57\% (1,449 WSIs) were collected from publicly available sources, including the Kidney Precision Medicine Project (KPMP)\cite{kpmp_data}, NEPTUNE~\cite{barisoni2013digital}, and HUBMAP~\cite{hubmap-kidney-segmentation,hubmap2019human}, while the remaining 43\% (1,093 WSIs) were acquired from an internal collection at Vanderbilt University Medical Center. To increase the diversity of our dataset, we incorporated WSIs from both human and rodent samples.

\textbf{Fine-tuning Dataset.} 
Building on the evaluation dataset construction described above, three foundation models were used for inference and curation. ``Good'' prediction masks and corresponding images from all three foundation models were collected as ``easy" image–pseudo-label pairs and added to the fine-tuning dataset. \textcolor{black}{Additionally, a small set of 198 hard image patches, which were consistently rated as ``bad” by student raters across all foundation models, were manually annotated by pathologists and included in the fine-tuning dataset.} As shown in Table~\ref{tab:Experiments},  \textcolor{black}{This resulted in a total of 12,005 image-label pairs. For the training and validation split, 11,807 pairs were used for training, while a diverse set of 100 pairs was reserved for validation. Lastly, an additional 185 images, sampled from all kidney WSI sources, were annotated by pathologists and designated as the hold-out testing set to evaluate the performance of the fine-tuned foundation models. Each 512 $\times$ 512 image patch contains 50 to 300 cell nuclei, providing substantial signals for effective fine-tuning. We ensure that no overlap existed between WSIs used for training/validation and those in the hold-out test set.} 

\subsubsection{Evaluation of Foundation Model Performance}

Nuclei instance segmentation results from the three foundation models were evaluated using the criteria shown in Fig.~\ref{fig:rating_details}. This rating system quantitatively assesses each model’s predictions on the evaluation dataset.

\textbf{Single Model Performance Evaluation.} First, we evaluated the performance of each individual foundation model by analyzing the distribution of rating assignments across our evaluation dataset. This analysis provided insights into each cell foundation model's performance and behavior by examining the frequency of prediction classes (``good," ``medium," ``bad") for each foundation model.

\textbf{Fused Model Performance Evaluation.} Building on the evaluation of each individual cell foundation model's performance, we conducted a joint model performance evaluation. Specifically, an image patch was assigned to the ``good" prediction class if any model rated it as ``good." Conversely, an image patch was categorized as ``bad" only if all foundation models were assigned a ``bad" rating. The remaining image patches were categorized as ``medium." This joint analysis indicates the upper bound of applying multiple foundation models to our domain-specific task and highlight the potential for fine-tuning these models through our data enrichment strategies. Furthermore, a taxonomy of common errors made by all cell foundation models in this study can be derived from the fused ``bad" image patches.

\textbf{Cross-Model Performance Evaluation.} In this work, we also evaluated cross-model performance on our kidney nuclei dataset by conducting an agreement analysis among the foundation models. First, we computed an \textbf{Agreement Matrix} that quantified the percentage of agreement between each pair of models, indicating how often they assigned the same rating to image patch predictions. This matrix highlights the consistency among the foundation models in their rating assignments. Next, we examined \textbf{Class-Specific Agreement} by categorizing image patches into three groups: \underline{All Three Models Agree}, where all models assigned the same rating; \underline{Two Models Agree}, where any two models assigned matching ratings; and \underline{No Agreement}, where none of the models assigned the same rating to the image patch.

\subsubsection{Data-Enriched Fine-Tuning with Multiple Foundation Models}

As described earlier, Each foundation model was fine-tuned using either ``easy" image patches (foundation model-generated pseudo labels), ``hard" image patches (pathologist-corrected), or both. Table~\ref{tab:Experiments} shows the details of the fine-tuning experiments. For each fine-tuning setting, we used different scales of the training dataset. Specifically, we selected common proportions—25\%, 50\%, 75\%, and 100\%—to evaluate whether the impact of annotation strategies across the three models remains consistent under different dataset sizes.

\vspace{10pt}
\begin{table*}[h]  
\centering
\small  
\begin{tabular}{|c|c|>{\centering\arraybackslash}p{1.5cm}|>{\centering\arraybackslash}p{1.5cm}|>{\centering\arraybackslash}p{1.5cm}|>{\centering\arraybackslash}p{1.5cm}|}  
\hline
\multirow{2}{*}{Fine-tuning Experiments} & \multirow{2}{*}{Data Labeling Source}            & \multicolumn{4}{c|}{Incremental Training Dataset Settings} \\ \cline{3-6} 
                                         &                                                  & 25\% & 50\% & 75\% & 100\%  \\ \hline
``Easy" patches only                      & Models' Predictions                              & 2,952 & 5,901 & 8,854 & 11,807 \\ \hline
``Hard" patches only                       & Pathologists' correction                         & 50    & 99    & 149   & 198    \\ \hline
``Easy" + ``Hard" patches                   & Combine Both                                   & 3,002 & 6,000 & 9,003 & 12,005 \\ \hline

\end{tabular}
\vspace{10pt}
\caption{The table summarizes the model fine-tuning experiments conducted under three settings: using \textbf{``easy"} image patches (foundation model-generated pseudo labels), \textbf{``hard"} image patches (pathologist-corrected), and a combination of both. Each data-enriched fine-tuning experiment was performed at different scales of the training dataset (25\%, 50\%, 75\%, and 100\%) for the three foundation models: Cellpose, StarDist, and CellViT. The numerical values indicate the size of the scaled training dataset for each individual fine-tuning experiment.}
\label{tab:Experiments}
\end{table*}

\subsubsection{Evaluation Metrics and Implementation Details}
\textbf{Evaluation metrics.} To assess the instance segmentation performance of the foundation models after fine-tuning, we used Recall, Precision, and F1 score as evaluation metrics. Details are provided in the Supplementary Experiments. 

\textbf{Implementations.} Both training and inference experiments were implemented using Python 3.9 and PyTorch 2.0.1, with GPU acceleration provided by CUDA 11.7. The experiments were conducted on an NVIDIA RTX A5000 GPU with 24GB of memory. For a simple implementation of model inference using each foundation model's released pretrained weights with GPU support, our previous work~\cite{guo2025assessment} provides customized object-oriented python modules. 

During fine-tuning, to ensure consistency in experimental settings and enable meaningful comparison, we applied the same oversampling factor \( \gamma_s = 0.85 \) across all experiments that used both ``easy" and ``hard" patches. Due to the imbalance between ``easy" and ``hard" samples, the number of training epochs was set to 50 for all fine-tuning experiments involving ``easy" image patches, while experiments using only ``hard" image patches were trained for 25 epochs. The batch size was consistently set to 16 across all experiments. Detailed information about model fine-tuning is provided in the Supplementary Experiments.

\subsection{Statistics and Reproducibility} 
Model evaluation was performed on 8,818 image patches sampled from 2,542 curated kidney whole slide images. Each patch was assigned a qualitative rating (``good,'' ``medium,'' or ``bad'') using established criteria. For inference with the publicly available Cellpose, StarDist, and CellViT models, we used our previously published object-oriented Python modules~\cite{guo2025assessment} to ensure consistent preprocessing and execution. To assessed the fine-tuned models, instance segmentation performance was assessed on the test dataset using Recall, Precision, and F1 score, calculated at the image-patch level with mean and standard deviation reported. Training parameters were standardized across models to enable fair comparison. Replicates refer to independent evaluations of each model on the same test dataset under identical inference conditions. All experiments were deterministic, with identical hardware to ensure reproducibility.

\subsection{Ethical Consideration}
This work was conducted with approval and oversight from the Vanderbilt University Institutional Review Board (IRB 230223). This is a retrospective study on archival tissue. No interventions and experimental manipulations were included. We did not recruit participants. We retrieved the archival kidney tissue biopsy samples from the tissue bank. We used pre-collected and de-identified microscopy data only without knowing the information (e.g., age, genotypic information, past and current diagnosis and treatment categories) of the population characteristics.

\section{Results}
\subsection{Evaluation of Foundation Model Performance}
\begin{figure*}[ht]
\begin{center}
\includegraphics[width=1\linewidth]{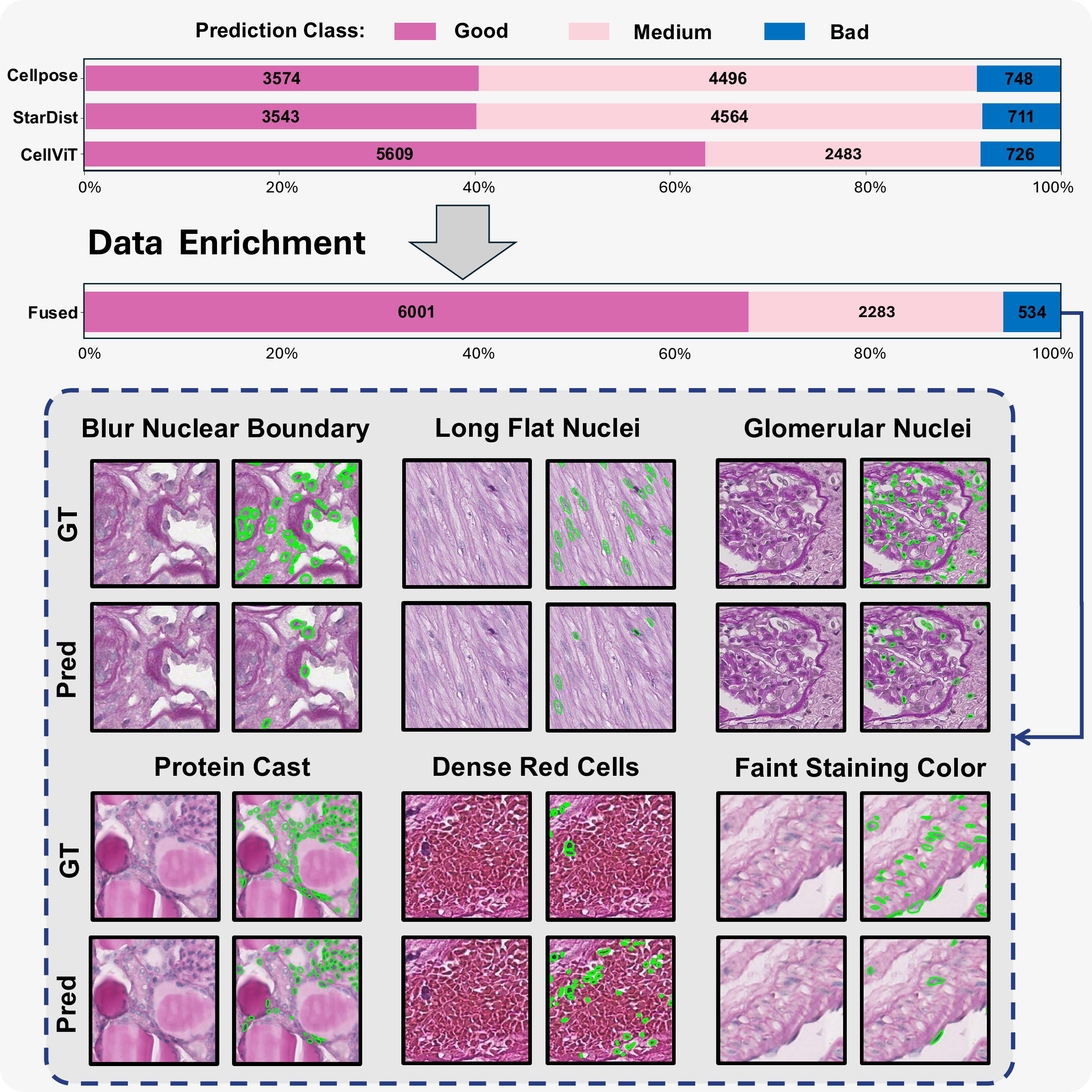}
\end{center}
\caption{\textbf{Distribution of Rated Predictions from Cell Foundation Models Across the Evaluation Dataset.} Each row represents the foundation model's predictions, with three values corresponding to the number of predictions rated as ``good," ``medium," and ``bad," respectively. Then, data enrichment (shown as "Fused" Model) was performed based on the evaluation results of individual models, resulting in an increase in ``good" image patches and a decrease in ``bad" image patches. Lastly, we summarized a taxonomy of ``bad" image patches that all foundation models failed.}
\label{fig:in_model}
\end{figure*}

\textcolor{black}{\textbf{Individual Model Performance Evaluation.} First, we rated the segmentation performance of each foundation model shown in the upper panel of Fig.~\ref{fig:in_model}. CellViT demonstrated superior performance in segmenting nuclei in these kidney images, with 5,609 (63.6\%) ``good" predictions, which outperforms Cellpose (40.5\%) and StarDist (40.2\%). This indicates that, despite not being specifically trained on kidney pathology nuclei data, the models retain some transferable knowledge applicable to our task. However, even with the best CellViT model, there is still an existence of approximately 37\% of rated predictions from ``medium" and ``bad," which underscores the need for a kidney domain-specific foundation model. Thus, the fused evaluation dataset—with enriched ``good” image patches (6,001; 68\%) and a reduced number of ``bad” patches (534; 6\%)— were further utilized for continuous fine-tuning.}

\textbf{Taxonomy of Model Failures.} As illustrated in the lower panel of Fig.~\ref{fig:in_model}, a taxonomy of common errors made by all cell foundation models is summarized from the fused set of ``bad" image patches. It is evident long and flat nuclei, nuclei with blurred boundaries, and densely distributed nuclei within glomeruli are particularly difficult to segment accurately. Additionally, slides with faint staining, those containing fatty tissues, or slides with an excessive number of red blood cells can result in a higher incidence of model failures.

\begin{figure*}[ht]
\begin{center}
\includegraphics[width=1\linewidth]{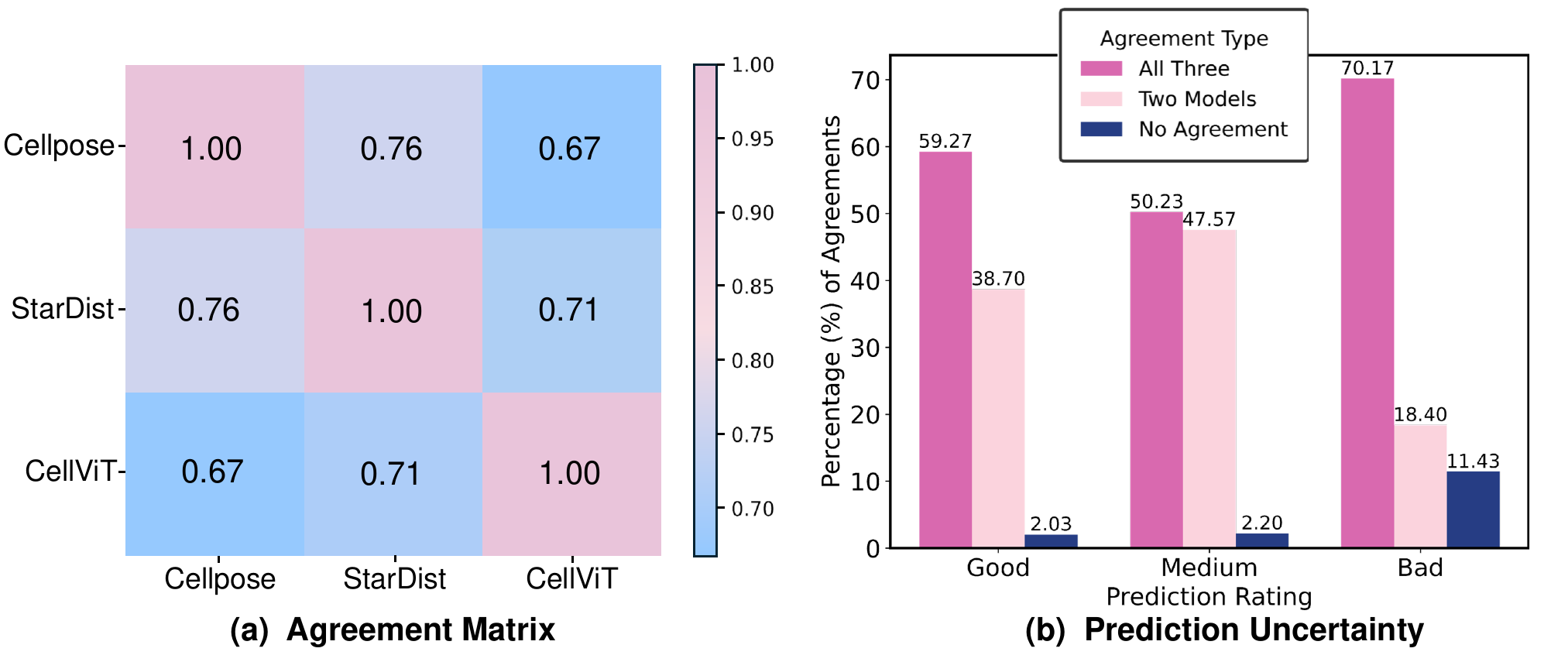}
\end{center}
\caption{\textbf{Cross-Model Performance Agreement.} (a) shows the agreement matrix between each pair of foundation models. To further assess the cross-model performance, (b) shows the percentages of image patches where \textbf{all three models agree}, \textbf{two models agree}, or \textbf{no models agree}, for each prediction class (“good”, “medium”, “bad”).}
\label{fig:cross_models}
\end{figure*}

\textbf{Cross-Model Performance Agreement.} Next, we assess their collective behavior by examining the agreement and disagreement among their predictions. Fig.~\ref{fig:cross_models}a shows the consistency between model predictions. As illustrated, no pair of models reaches over 90\% agreement or complete disagreement. The highest agreement occurs between Cellpose and StarDist (0.76), while the lowest is between Cellpose and CellViT (0.67). These non-extreme values, ranging from 0.67 to 0.76, suggest that the current SOTA nuclei foundation models in digital pathology do not generalize exceptionally well to large-scale, diverse kidney datasets. Each model exhibits its own strengths, highlighting the potential for combining multiple SOTA foundation models to improve performance in downstream kidney nuclei tasks. Fig.~\ref{fig:cross_models}b shows the percentages of image patches where all three models agree, two models agree, or no models agree for each rated prediction class (``good", ``medium", ``bad"). It is evident that ``bad" samples exhibit the highest inter-model reliability, with over 70\% agreement among all three models, indicating the nature of our design that prioritizes human annotation effort for consensus ``bad" image patches.

\textbf{Inter-Rater Agreement.} \textcolor{black}{Lastly, we also randomly selected 300 images to assess inter-rater reliability, as shown in Supplementary Table S1. CellViT, which has the strongest baseline performance, achieved the highest agreement of 94.3\% between the two raters.}

\begin{table}[ht]
\centering
\begin{minipage}{0.48\textwidth} 
    \centering
    \resizebox{\linewidth}{!}{
    \begin{tabular}{lccc}
    \toprule
    \textbf{Fine-tuned Cellpose Model} & \textbf{F1 score} & \textbf{Precision} & \textbf{Recall}
    \\
    \addlinespace[0.7em]
     Cellpose Baseline & \textcolor{blue}{0.6748} & \textcolor{blue}{0.7006} & \textcolor{blue}{0.6694} \\ 
     \midrule
    Easy 10\%  & 0.7038 & 0.8272 & 0.6276 \\ 
    Easy 25\%  & 0.7297 & 0.8313 & 0.6649 \\ 
    Easy 50\%  & 0.7357 & 0.8286 & 0.6748 \\ 
    Easy 75\%  & 0.7422 & 0.8447 & 0.6747 \\ 
    Easy 100\% & \textcolor{red}{0.7529} & 0.8349 & \textcolor{red}{0.6986} \\ 
    \midrule
    Hard 10\%   & 0.5117 & 0.8236 & 0.3905 \\
    Hard 25\%   & 0.5353 & 0.8095 & 0.4170 \\ 
    Hard 50\%   & 0.6444 & 0.7959 & 0.5592 \\ 
    Hard 75\%   & 0.5671 & 0.8401 & 0.4461 \\ 
    Hard 100\%  & 0.5807 & 0.8121 & 0.4699 \\ 
    \midrule
    Easy 10\% + Hard 10\%   & 0.7067 & 0.8444 & 0.6237 \\ 
    Easy 25\% + Hard 25\%   & 0.7228 & 0.8475 & 0.6466 \\ 
    Easy 50\% + Hard 50\%   & 0.7270 & 0.8481 & 0.6493 \\ 
    Easy 75\% + Hard 75\%   & 0.7352 & \textcolor{red}{0.8535} & 0.6953 \\ 
    Easy 100\% + Hard 100\% & 0.7350 & 0.8512 & 0.6601 \\ 
    \bottomrule
    \end{tabular}
    }
    \vspace{10pt}

\end{minipage}%
\hfill  
\begin{minipage}{0.48\textwidth}  
    \centering
    \resizebox{\linewidth}{!}{
    \begin{tabular}{lccc}
    \toprule
    \textbf{Fine-tuned StarDist Model} & \textbf{F1 score} & \textbf{Precision} & \textbf{Recall} \\ 
    \addlinespace[0.7em]
   StarDist Baseline & \textcolor{blue}{0.7380} & \textcolor{blue}{0.9158} & \textcolor{blue}{0.6331} \\ 
    \midrule
    Easy 10\%  & 0.7838 & 0.8962 & 0.7113 \\ 
    Easy 25\%  & 0.7926 & 0.8989 & 0.7209 \\ 
    Easy 50\%  & 0.7931 & \textcolor{red}{0.9084} & 0.7166 \\ 
    Easy 75\%  & 0.7895 & 0.9063 & 0.7110 \\ 
    Easy 100\% & 0.7943 & 0.8991 & 0.7234 \\ 
    \midrule
    Hard 10\%   & 0.7876 & 0.8577 & 0.7435 \\ 
    Hard 25\%   & 0.8086 & 0.8862 & 0.7423 \\ 
    Hard 50\%   & 0.8166 & 0.8834 & 0.7696 \\ 
    Hard 75\%   & 0.8109 & 0.8864 & 0.7601 \\ 
    Hard 100\%  & \textcolor{red}{0.8229} & 0.8847 & \textcolor{red}{0.7802} \\ 
    \midrule
    Easy 10\% + Hard 10\%   & 0.7899 & 0.9022 & 0.7152 \\ 
    Easy 25\% + Hard 25\%   & 0.7980 & 0.9017 & 0.7249 \\ 
    Easy 50\% + Hard 50\%   & 0.8110 & 0.9015 & 0.7464 \\ 
    Easy 75\% + Hard 75\%   & 0.8172 & 0.8961 & 0.7588 \\ 
    Easy 100\% + Hard 100\% & 0.8182 & 0.8901 & 0.7664 \\ 
    \bottomrule
    \end{tabular}
    }
    \vspace{10pt}

\end{minipage}

\vspace{20pt}  

\begin{minipage}{0.48\textwidth}  
    \centering
    \resizebox{\linewidth}{!}{
    \begin{tabular}{lccc}
    \toprule
    \textbf{Fine-tuned CellViT Model} & \textbf{F1 score} & \textbf{Precision} & \textbf{Recall} \\ 
    \addlinespace[0.7em]
    CellViT Baseline & \textcolor{blue}{0.7838} &\textcolor{blue}{0.8233} & \textcolor{blue}{0.7629} \\ 
    \midrule
    Easy 10\%  & 0.7687 & 0.8287 & 0.7276 \\ 
    Easy 25\%  & 0.7836 & 0.8216 & 0.7621 \\ 
    Easy 50\%  & 0.7828 & 0.8296 & 0.7520 \\ 
    Easy 75\%  & 0.7854 & 0.8284 & 0.7587 \\ 
    Easy 100\% & 0.7794 & 0.8270 & 0.7476 \\ 
    \midrule
    Hard 10\%   & 0.7397 & 0.7707 & 0.7295 \\ 
    Hard 25\%   & 0.7796 & 0.7866 & 0.7857 \\ 
    Hard 50\%   & 0.7770 & 0.7898 & 0.7791 \\ 
    Hard 75\%   & 0.7881 & 0.7770 & \textcolor{red}{0.8137} \\ 
    Hard 100\%  & 0.7861 & 0.7809 & 0.8070 \\ 
    \midrule
    Easy 10\% + Hard 10\%   & 0.7791 & 0.8216 & 0.7498 \\ 
    Easy 25\% + Hard 25\%   & 0.7893 & 0.8259 & 0.7695 \\ 
    Easy 50\% + Hard 50\%   & 0.7895 & 0.8276 & 0.7652 \\ 
    Easy 75\% + Hard 75\%   & 0.7866 & \textcolor{red}{0.8303} & 0.7580 \\ 
    Easy 100\% + Hard 100\% & \textcolor{red}{0.7952} & 0.8302 & 0.7731 \\ 
    \bottomrule
    \end{tabular}
    }
    \vspace{10pt}

\end{minipage}%
\hfill  
\caption{\textbf{Fine-tuning Results for Cellpose, StarDist, and CellViT Models.} This table consists of three subtables summarizing the instance segmentation performance metrics (F1 score, Precision, and Recall) for each foundation model across three data enrichment experimental settings (using only ``easy" images, only pathologist-corrected ``hard" images, and a combination of both). Each sub-table compares the baseline performance (highlighted in \textcolor{blue}{blue}) of the model against its fine-tuned counterparts. The highest values of each evaluation metric for the fine-tuned models are highlighted in \textcolor{red}{red}.}
\label{tab:finetuning_results}
\end{table}

\begin{figure*}[hb]
\begin{center}
\includegraphics[width=1\linewidth]{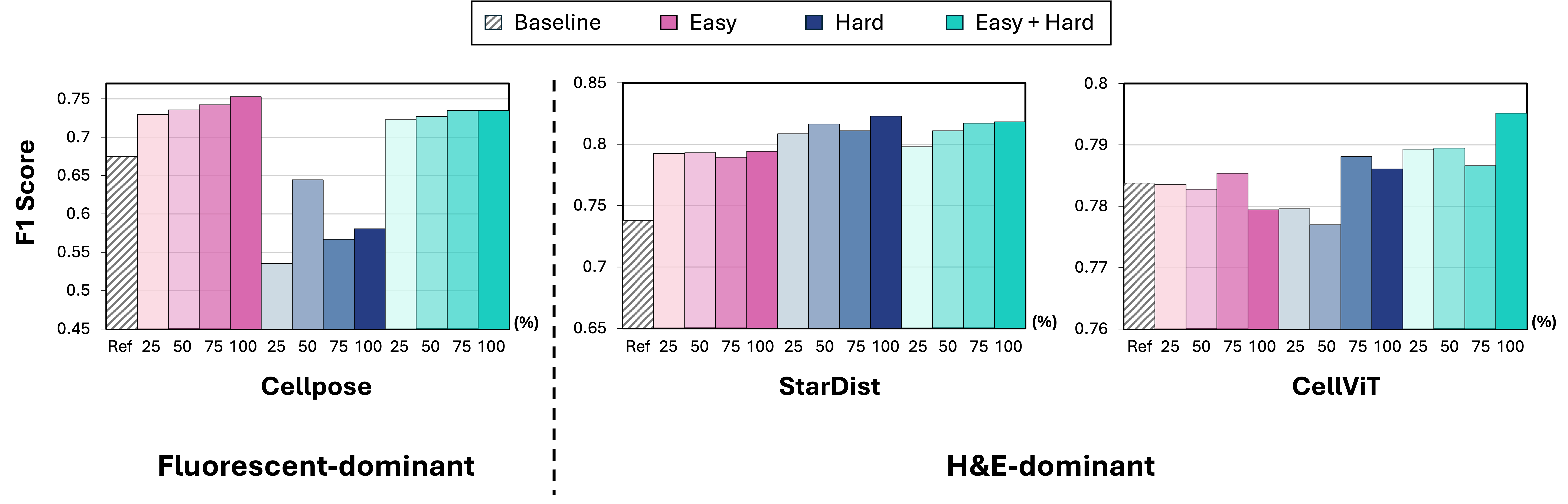}
\end{center}
\caption{\textcolor{black}{\textbf{Comparison of F1-score across training/annotation strategies.} The horizontal axis represents the percentage (\%) of labels included under different strategies. Cellpose achieves optimal performance with ``easy” labels, StarDist with ``hard” labels, and CellViT with both. In this context, ``hard” cases refer to training instances where automatic segmentation fails human quality assurance (QA) and employ pixel-level corrections, whereas ``easy” cases are those where automatic results pass human QA.}}
\label{fig:finetuning_best}
\end{figure*}

\subsection{Data-Enriched Foundation Models Fine-Tuning}
\textbf{Baseline Performance.} First, We used the pretrained weights of each foundation model as the baseline and evaluated their performance on our hold-out test set; results are shown in blue in Table~\ref{tab:finetuning_results}. Consistent with prior human feedback, CellViT outperforms the other two models, with an F1 score of 0.7838, followed by StarDist at 0.7380, and Cellpose at 0.6748. Although StarDist achieves a high precision score of 0.9158, its lower recall of 0.6331 suggests that it misses certain amount of kidney cell nuclei. Cellpose exhibits the largest domain gap with the lowest F1 score and shows at least 10\% decrease in all evaluation metrics compared to CellViT.

\textbf{Fine-tuned Performance.} Then, we investigated the performance of foundation models from our data-enriched fine-tuning strategies. The details of instance segmentation performance gains are provided below with the evaluation metric values referenced in Table~\ref{tab:finetuning_results}.

\textbf{(1) Cellpose:} As shown, training on ``hard" patches alone significantly degrades performance, particularly in the F1 score and Recall. In contrast, the inclusion of large-scale ``easy" (pseudo-labeled) training samples consistently improves performance, with optimal results achieved using only ``easy” patches, significantly boosting all metrics over the baseline (F1: 0.6748 $\rightarrow$ 0.7529).

\textbf{(2) StarDist:} All fine-tuning strategies lead to strong instance segmentation performance gains, especially in F1 (0.738 $\rightarrow$ 0.8229) and Recall (0.6331 $\rightarrow$ 0.7802). While Precision saw a slight drop from 0.9158 to around 0.89, the decrease is minimal compared to the substantial improvements in F1 and Recall.  ``Easy” patches support knowledge distillation, while ``hard” ones close domain gaps, confirming the effectiveness of HITL-guided data enrichment.

\textbf{(3) CellViT:} Performance remains stable across strategies. Combining both ``easy” and ``hard” patches yields the highest F1 score (0.7952) and the most balanced results, though the gains are smaller than those observed for Cellpose or StarDist. This may be due to noise introduced by pseudo-labels from less accurate models, which will be examined in the next section~\ref{subsec:label_noise}.

\textcolor{black}{\textbf{Optimal Annotation Strategies.} Lastly, we summarize the optimal training and annotation strategies for the three foundation models in Fig.~\ref{fig:finetuning_best}. We use the F1-score as the primary evaluation metric, as it balances both precision and recall. Among the models, the \underline{\textit{Fluorescent-dominant}} model, \textbf{Cellpose}, achieves optimal performance when trained with only ``easy” labels. Its performance degrades with the inclusion of ``hard” labels, likely due to their significant domain difference from the model's pre-training data. In this context, ``hard” cases are the training cases where automatic segmentation fails human QA and employ pixel-level manual corrections, while ``easy” cases are those where the automatic results successfully pass human QA without modification. In contrast, the \underline{\textit{H\&E-dominant}} models consistently benefit from incorporating human-corrected labels (``hard”), outperforming their respective baselines. Specifically, \textbf{StarDist} achieves the best performance with ``hard” labels (attaining the highest F1 score of 0.8229), while \textbf{CellViT} performs optimally when trained with both ``easy” and “hard” labels. In summary, the combined use of ``easy" and ``hard" patches proved to be the most effective and broadly applicable annotation strategy, consistently enhancing performance across all three models.}

\textcolor{black}{\textbf{Choice of training dataset size.}} In this study, we selected common proportions—25\%, 50\%, 75\%, and 100\%—to evaluate whether the impact of annotation strategies across the three models remains consistent under different dataset sizes. Given that the annotation efforts are limited to only a total of 198 ``hard” labels and a majority comes from pseudo-labels. We hypothesize that maintaining the overall scale of the training dataset to include enough representative samples is important. To further assess this, we conducted additional experiments using only \textbf{10\%} of ``easy” labels (N = 1,181), ``hard” labels (N = 20), and their combination to validate the results. The outcomes are also presented in Table \ref{tab:finetuning_results}. 

As shown in Table \ref{tab:finetuning_results}, reducing the training dataset to 10\% results in smaller performance gains across all models, though the overall trend remains consistent—particularly for the CNN-based models (Cellpose and StarDist), as illustrated in Fig. \ref{fig:finetuning_best}. In contrast, performance drops more noticeably below the baseline for the ViT-based CellViT. With only 20 ``hard” samples, fine-tuning becomes challenging, making comparisons between groups (e.g., ``easy” vs. ``easy” + ``hard”) less reliable. To broadly improve model performance, a sufficient amount of both ``hard” and ``easy” data (e.g., at least 25\%) is essential for effective fine-tuning. Alternatively, maintaining segmentation accuracy with smaller datasets requires additional data curation to ensure representative coverage.

\textcolor{black}{\textbf{Robustness.} To better evaluate the robustness of the experiments, we applied 5-fold cross-validation and compared F1-scores as part of an ablation study. As shown in Supplementary Table S2, the results compare model performance under different training annotation strategies.}

\subsection{Effect of Label Imbalance on Model Performance}
\label{subsec:label_noise}
\begin{figure*}[ht]
\begin{center}
\includegraphics[width=1\linewidth]{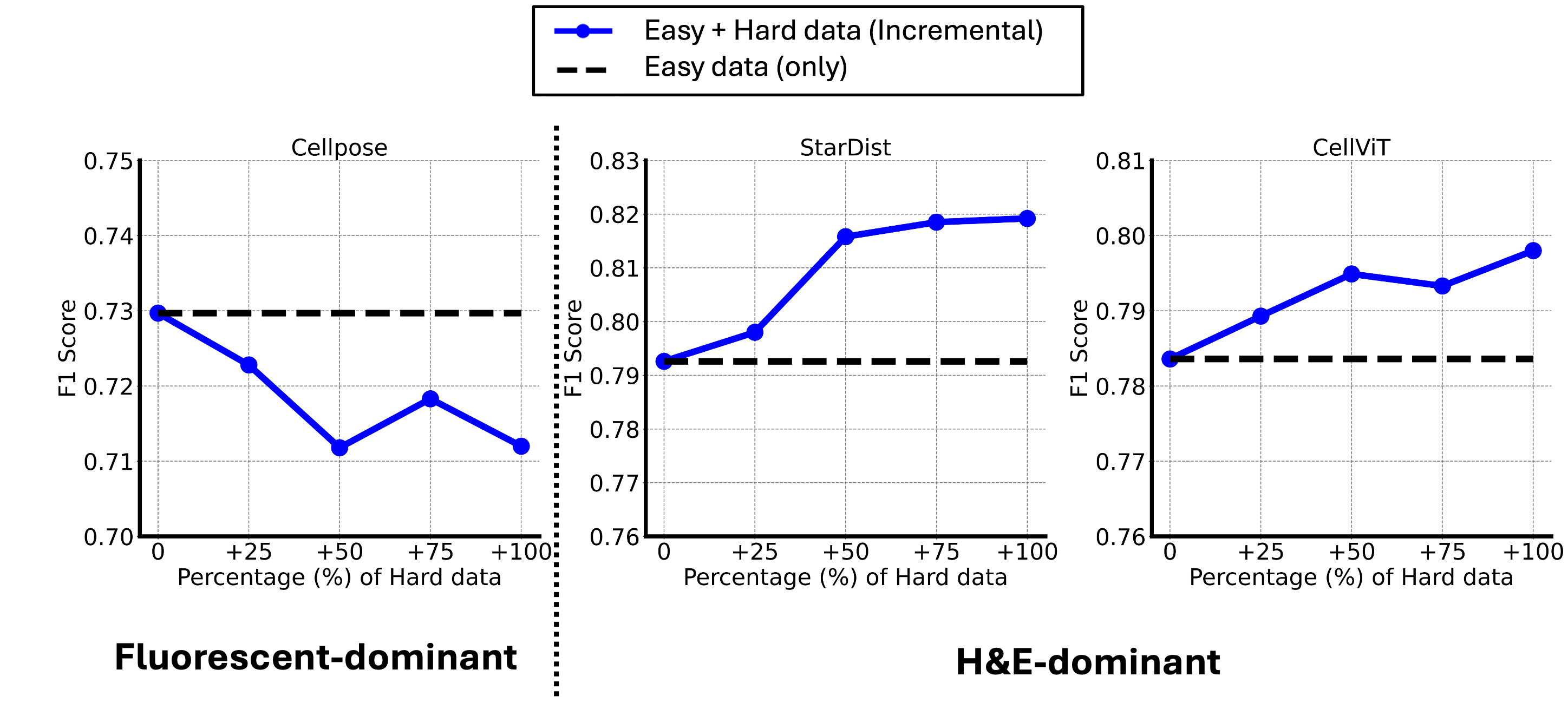}
\end{center}
\caption{\textcolor{black}{\textbf{Impact of Incrementally Increasing Percentage (\%) of  Noise-Free, Pathologist-Corrected ``Hard" Labeled Data on Model Performance.} As shown, the inclusion of ``hard" labeled images decreases the performance of Cellpose. In contrast, for both CellViT and StarDist, incorporating these data leads to increased F1 scores.}}
\label{fig:bad_incremental}
\end{figure*}

\textcolor{black}{Compared to Cellpose and StarDist, CellViT shows a smaller performance gain over its baseline, which we hypothesize is due to \underline{imbalanced annotation quality}. \textbf{(1) variation in pseudo-label quality introduces noise.} As shown in Fig.~\ref{fig:in_model}, CellViT—the strongest baseline model—generates the majority of pseudo labels; however, incorporating pseudo labels from weaker models may introduce noise that hinders its fine-tuning. \textbf{(2) There is an imbalance between pseudo labels (from ``easy'' cases) and gold-standard annotations (from ``hard'' cases).} To investigate these effects, we conducted an ablation study in which we incrementally added noise-free, pathologist-corrected ``hard'' image patches to the training set, as shown in Fig.~\ref{fig:bad_incremental}.} 

\textcolor{black}{Consistent with Fig.~\ref{fig:finetuning_best}, the Fluorescent-dominant Cellpose model shows decreased performance when ``hard'' patches are included, although the annotations are gold-standard, but the usually faintly stained patches are too different from its pre-training dataset. In contrast, the H\&E-dominant models, CellViT and StarDist, show consistent performance improvement as more high-quality ``hard'' annotations are added. The differing trend may be attributed to the training data used for each model: the Cellpose model was predominantly trained on fluorescence microscopy images with Fluorescent staining (a ``Fluorescent-dominant” model), while CellViT and StarDist were primarily trained on histology images with H\&E staining (forming ``H\&E-dominant” models). Since the “hard” dataset is composed mostly of H\&E-stained images, this domain gap likely contributes to the observed differences in performance. Notably, CellViT achieves its highest F1 score of 0.798 when all ``hard” image patches are included. This aligns with the optimal annotation strategies shown in Fig.~\ref{fig:finetuning_best}, where the combined use of ``easy” (pseudo-labeled) and ``hard” (pathologist-corrected) patches proves to be the most effective and broadly applicable approach, consistently improving performance across all three models.}

\begin{figure*}[ht]
\begin{center}
\includegraphics[width=0.9\linewidth]{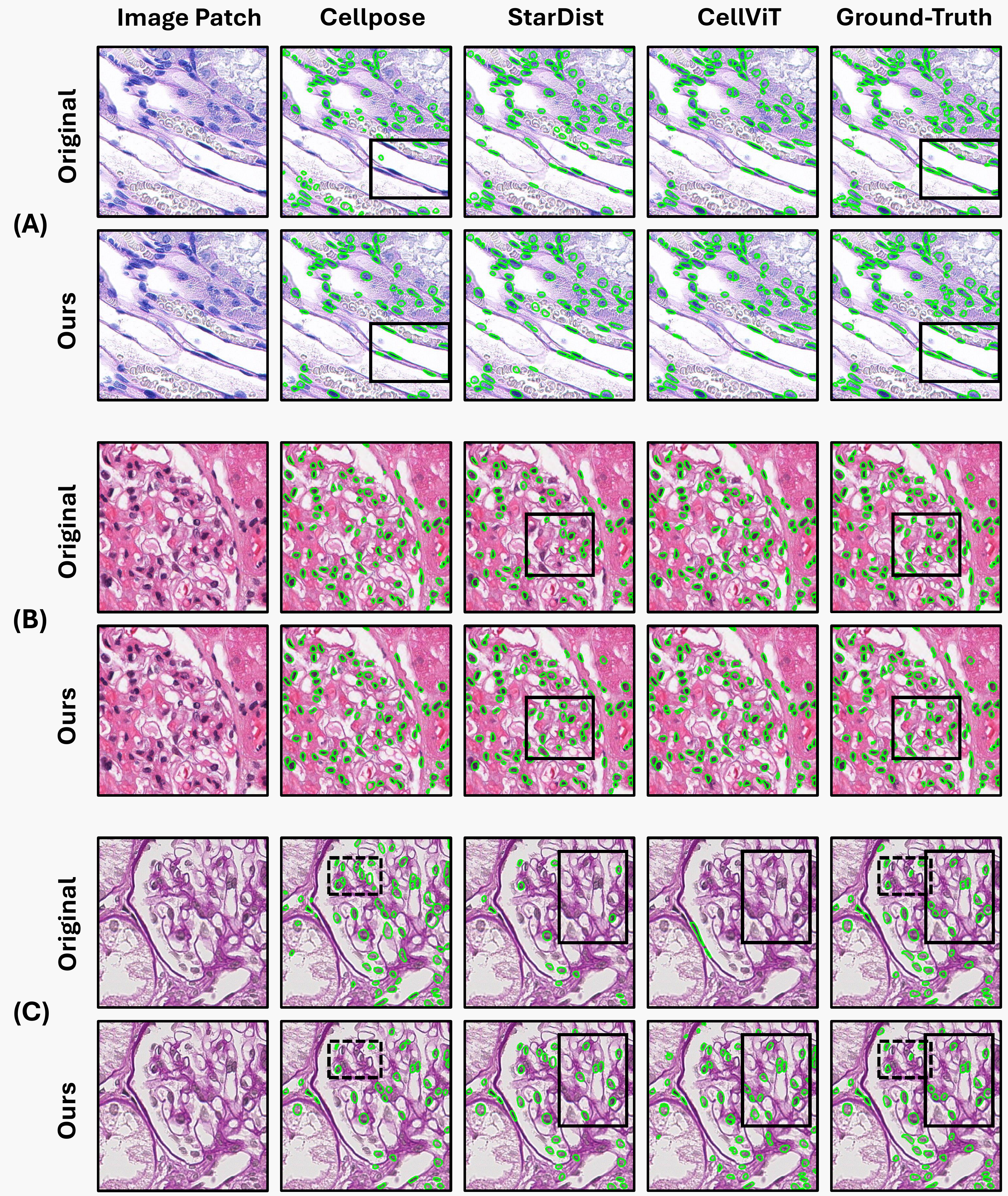}
\end{center}
\caption{\textbf{Qualitative Results of Enhanced Kidney Cell Nuclei Instance Segmentation.} The predicted nuclei and ground truth are represented as overlaid green contours on the image patch, with areas of improvement highlighted by rectangles.}
\label{fig:performance_qualitative}
\end{figure*}

\subsection{Visualization of Improved Model Performance}

In this section, we present the qualitative visualization results of our enhanced kidney cell nuclei instance segmentation, evaluated using the best fine-tuning strategies from three foundation models. Three example image patches (A, B, and C) are presented in Fig.~\ref{fig:performance_qualitative}. The upper panel for each example shows the baseline performance across all models, while the lower panel displays the segmentation performance following data-enriched fine-tuning. As shown in Fig.~\ref{fig:performance_qualitative}(A), our fine-tuned Cellpose model improves the instance segmentation on long and flat nuclei. (B) highlights the improvement of our StarDist model in segmenting dense nuclei in a glomerulus. The last panel shows the improvement in segmenting nuclei in PAS-stained images, which are underrepresented in the current histology dataset. Most faintly stained nuclei that were missing in the original predictions within a glomerulus are detected by the fine-tuned StarDist and CellViT models. In contrast, the fine-tuned Cellpose model generates fewer false positive predictions compared to its baseline. Additionally, as illustrated in both (A) and (B), the highlighted regions in the image patches represent knowledge gaps in the target domain where the inferior model struggles with certain types of image patches, while the other models do not. Examples of our fine-tuning results demonstrate that this less effective model can benefit from the knowledge of the other models to improve its performance. Additional visualizations of the qualitative results (D)-(L) are provided in Supplementary Figs. S1-S3.

\section{Discussion}
\textcolor{black}{The evaluation of Cellpose, StarDist, and CellViT reveals a persistent performance gap between general and kidney-specific nuclei segmentation, indicating that domain-specific tuning is still required.} Our foundation model-based data enrichment strategies aim to meet the need for a kidney-level cell foundation model and to reduce labeling costs. In this process, the collective image curation from multiple foundation models is crucial, as a single model may not capture the diversity of segmentation styles in the target kidney domain. By leveraging the versatility of multiple foundation models, annotation efforts are minimized and transformed into time-efficient image rating curation, which can be further enhanced through the model's uncertainty analysis. In the first round of HITL fine-tuning, all three foundation models demonstrated improved performance over their baselines, with StarDist achieving the highest segmentation performance, achieving an F1 score of 0.82. The combined use of ``easy" (pseudo-labeled) and ``hard" (pathologist-corrected) patches proved to be the most effective and broadly applicable strategy, consistently improving performance across all three models. Although Cellpose shows a significant increase in performance (with the F1 score rising from 0.67 to 0.75) with our fine-tuning, it demonstrates lower efficacy when only hard image samples are included. This is primarily because the Cellpose model is primarily designed for fluorescently labeled images (Fluorescent-dominant model). For the training of segmentation task, 512 $\times$ 512 images were converted into grayscale nuclear bands, which can lead to information loss, particularly for hard images with very light staining colors and complex tissue morphological patterns. In contrast, the other two models have been adapted for H\&E-stained images. 

To clarify the definition of pathology-related foundation models in our study, the three models are image-only foundation models, highlighting scale, data modalities (incorporating different microscopy techniques). Among three models, CellViT is a transformer-based model built upon DINO/SAM and regarded as a foundation model in recent works~\cite{bilal2025foundation, lee2024foundation, israel2023foundation}. Cellpose and StarDist are CNN-based generalist models. CNN-based models that are pre-trained with large-scale representative data are also regarded as foundation models in~\cite{pai2023foundation, he2024vista3d}.

While applying multiple foundation models leverages each model’s specialty and reduces annotation costs, our data enrichment and continual learning strategy still heavily relies on pseudo-labels, a weakly supervised approach. Thus, label quality may introduce noise. As shown in Table \ref{tab:finetuning_results} and Fig. \ref{fig:finetuning_best}, baseline model performance varies. For instance, although both the Cellpose and StarDist models contribute a similar number of ``easy” (i.e., ``good”) patches, the inferior baseline model (e.g., Cellpose for our kidney images) might produce lower-quality pseudo-labels compared to others. Additionally, since the image-level rating category does not fully capture local performance, it cannot completely reflect mask quality. These factors suggest that disparities can introduce noise into the fine-tuning dataset.

\textcolor{black}{Biases can be introduced by the rating system. (1) \textbf{Limited rating categories }(``good”, ``medium”, ``bad”). We quantify the percentage of correctly predicted nuclei per category, yet the ``medium” class often exhibits the highest variance, as demonstrated in Fig. \ref{fig:cross_models}'s cross-model evaluation. This ambiguity can lead to inconsistent ratings and uncertainty. (2) \textbf{Inter-rater variability} may stem from subjective observation practices, particularly since nuclei are small and their boundaries are indistinct. Even with a pathologist's final review, these biases cannot be entirely eliminated. (3) \textbf{Image-level ratings may overlook fine-grained details in patch predictions.} Therefore, introducing local-level ratings, along with improved rating categories, could enhance the system’s generalizability and robustness.}

As the models become better adapted to the specific kidney organ and the performance gap decreases among three models after the first round of fine-tuning, we anticipate generating improved curation results, including a higher number of ``good" ratings and more accurate prediction masks in the next round of HITL fine-tuning. Additionally, improved accuracy in nuclei instance segmentation is critical for the extraction and reliability of cell pathomic features (e.g., size, shape, area). As in our previous studies~\cite{deng2024hats, deng2025casc}, the future work will involve implementing a joint training scheme that combines classification with segmentation, enabling the classification task to guide the segmentation process.

In conclusion, our evaluation of a comprehensive kidney dataset shows that cell nuclei segmentation in histopathology requires further improvement through more organ-targeted foundation models (e.g., kidney-specific). The proposed human-in-the-loop data enrichment strategy—combining predictions from multiple models with limited expert corrections on challenging cases—consistently enhances performance across all models. Improved organ-specific foundation models can further increase the accuracy and reliability of cell pathomic feature extraction (e.g., size, shape, area) and can be integrated into QuPath~\cite{bankhead2017qupath} software to support more efficient workflows in clinical kidney pathology.

\section{Data availability}
The source data that support the plots within this manuscript is included in the Supplementary Data. The raw data is available upon reasonable request to the corresponding author [Y.H.] (yuankai.huo@vanderbilt.edu) and after satisfying Vanderbilt University and Vanderbilt University Medical Center's data use agreement.

\section{Code availability}
Code will be made publicly available at the AFM\_kidney\_cells repository~\cite{afm_kidney_cells}.

\section{Supplementary Information}
Supplementary materials are provided in the AFM\_kidney\_cells repository~\cite{afm_kidney_cells} within the ``stage2\_paper" directory.


\bibliography{main}   
\bibliographystyle{spiejour}   

\section* {Acknowledgments}
This work was supported by the National Institutes of Health under award numbers R01EB017230, R01DK135597, T32EB001628, K01AG073584, and 5T32GM007347, DoD HT9425-23-1-0003 (HCY), and in part by the National Center for Research Resources and Grant UL1 RR024975-01. This study was also supported by the National Science Foundation (1452485, 1660816, and 1750213). The Vanderbilt Institute for Clinical and Translational Research (VICTR) is funded by the National Center for Advancing Translational Sciences (NCATS) Clinical Translational Science Award (CTSA) Program, Award Number 5UL1TR002243- 03. The content is solely the responsibility of the authors and does not necessarily represent the official views of the NIH or NSF. This work was also supported by Vanderbilt Seed Success Grant, Vanderbilt Discovery Grant, and VISE Seed Grant. We extend gratitude to NVIDIA for their support by means of the NVIDIA hardware grant. This work was also supported by NSF NAIRR Pilot Award NAIRR240055. 

The KPMP is funded by the following grants from the NIDDK: U01DK133081, U01DK133091, U01DK133092, U01DK133093, U01DK133095, U01DK133097, U01DK114866, U01DK114908, U01DK133090, U01DK133113, U01DK133766, U01DK133768, U01DK114907, U01DK114920, U01DK114923, U01DK114933, U24DK114886, UH3DK114926, UH3DK114861, UH3DK114915, UH3DK114937. The content is solely the responsibility of the authors and does not necessarily represent the official views of the National Institutes of Health. The results here are in whole or part based upon data generated by the HuBMAP Program: https://hubmapconsortium.org. The Nephrotic Syndrome Study Network Consortium (NEPTUNE), U54-DK-083912, is a part of the National Institutes of Health (NIH) Rare Disease Clinical Research Network (RDCRN), supported through a collaboration between the Office of Rare Diseases Research (ORDR), NCATS, and the National Institute of Diabetes, Digestive, and Kidney Diseases. Additional funding and/or programmatic support for this project has also been provided by the University of Michigan, the NephCure Kidney International and the Halpin Foundation. The views expressed in written materials or publications do not necessarily reflect the official policies of the Department of Health and Human Services; nor does mention by trade names, commercial practices, or organizations imply endorsement by the U.S. Government.

\section*{Author Contributions}
Conceptualization: J.G., C.C., H.Y., Y.H.;
Study design: J.G., R.D., H.Y., Y.H.;
Investigation: J.G., C.C., Y.H.;
Data curation and annotation: J.G., S.L., Z.T., Y.L., M.Y., H.Y., Y.H.;
Visualization: J.G., S.L., R.D., H.Y., Y.H.;
Writing: J.G., S.L., H.Y., Y.H.;
Review and approve final manuscript: S.L., C.C., R.D., T.Y., M.L., Q.L., J.X., Y.W., S.Z., C.C. (Catie Chang), M.W., A.F., M.Y., H.Y., Y.H.; Funding acquisition: H.Y., Y.H.; Supervision: Y.H.;

Correspondence to Yuankai Huo (yuankai.huo@vanderbilt.edu)

\end{spacing}
\end{sloppypar}
\end{document}